%% file: m3imic.tex
\documentclass[lettersize,journal]{IEEEtran}
\usepackage{amsmath,amsfonts}
\usepackage{algorithmic}
\usepackage{algorithm}
\usepackage{array}
\usepackage[caption=false,font=normalsize,labelfont=sf,textfont=sf]{subfig}
\usepackage{textcomp}
\usepackage{stfloats}
\usepackage{url}
\usepackage{verbatim}
\usepackage{graphicx}
\usepackage{cite}

\usepackage{booktabs}
\usepackage{multirow}
\usepackage{subcaption}
\usepackage{threeparttable}
\usepackage{xcolor}

\newcommand{\oursmethod}{\ensuremath{\text{M3imic}}}

\begin{document}

\title{M3imic: Learning a Versatile Whole-Body Controller for Multimodal Motion Mimicking}


\author{
Zuxing~Lu,
Ziang~Zheng,
Yao~Lyu,
Jingyu~Liu,
Feihong~Zhang,
Song~Lu,
Xin~Yuan,
Changyin~Sun,~\IEEEmembership{Senior Member,~IEEE},
Xingxing~Zuo,~\IEEEmembership{Member,~IEEE},
and~Shengbo~Eben~Li,~\IEEEmembership{Senior Member,~IEEE},
\thanks{Manuscript received February 26, 2026; revised May 22, 2026; accepted June 1, 2026.
This work was supported in part by the National Natural Science Foundation of China under Grant (62203113), 
Tsinghua-Efort Joint Research Center for EAI Computation and Perception and SunRisingAI Lab,
 (Corresponding authors: Changyin~Sun (cysun@seu.edu.cn); Xingxing~Zuo (xingxing.zuo@mbzuai.ac.ae); Shengbo~Eben~Li (lishbo@tsinghua.edu.cn).)
}
\thanks{Zuxing~Lu, Jingyu~Liu, Xin~Yuan and Changyin~Sun are with the School of Automation, Southeast University, Nanjing 210096, China.
Ziang~Zheng, Yao~Lyu, Feihong~Zhang, Song~Lu and Shengbo~Eben~Li are with the School of Vehicle and Mobility, Tsinghua University, Beijing 100084, China.
Xingxing~Zuo is with the Department of Robotics, Mohamed Bin Zayed University of Artificial Intelligence, Abu Dhabi 7909, UAE.
}
}

\markboth{Journal of \LaTeX\ Class Files,~Vol.~14, No.~8, August~2021}%
{Shell \MakeLowercase{\textit{et al.}}: A Sample Article Using IEEEtran.cls for IEEE Journals}

\IEEEpubid{0000--0000/00\$00.00~\copyright~2021 IEEE}

\maketitle


\begin{abstract}
Building a general-purpose whole-body controller is essential for enabling diverse motion capabilities in humanoid robots across a wide range of downstream tasks, including locomotion and loco-manipulation.
Different tasks rely on distinct motion reference modalities: locomotion primarily depends on coordinated robot joint trajectories, whereas manipulation requires precise end-effector trajectory tracking.
Existing methods often overlook the representational mismatch between dense robot joint angles and sparse end-effector poses.
To address this, we propose Multi-Modal Mimic (M3imic), a versatile multi-modal whole-body control framework that unifies heterogeneous motion reference modalities, including robot joint angles, human pose trajectories, and end-effector poses, using modality-specific encoders to map them into a shared latent space.
Leveraging large-scale reinforcement learning in the simulator, we train a single policy that achieves sim-to-real transfer across multiple motion reference modalities without modality-specific retraining.
Extensive simulation and real-world experiments on the Unitree G1 robot are conducted to evaluate the proposed framework. In simulation, the policy achieves a peak success rate of 98.42\% on an unseen test dataset, demonstrating its exceptional generalization capability.
The code is available at \url{https://github.com/Renforce-Dynamics/MultiModalWBC}
\end{abstract}



\input{section/S1_intro}
\input{section/S2_relate_work}
\input{section/S3_method}

\input{section/S4_experiments}

\input{section/S5_conclusion}

\bibliographystyle{IEEEtran}
\bibliography{reference.bib}


\end{document}

%% file: section/S1_intro.tex
\section{Introduction}

\IEEEPARstart{I}{n} recent years, data-driven reinforcement learning~\cite{li2023reinforcement} has become an increasingly popular direction in humanoid robotics, complementing established model-driven control methodologies~\cite{schwenzer2021review, al2016multi}.
The ultimate goal of humanoid robots is to emulate human behaviors through anthropomorphic morphology, thereby assisting or augmenting humans across diverse tasks. 
Aiming for this, recent research has increasingly focused on enabling human-like control.
Progress in this direction has been facilitated, by the development of large-scale parallel simulation platforms~\cite{NVIDIA_Isaac_Sim} and advances in deep reinforcement learning algorithms, particularly in motion stylization~\cite{2021-TOG-AMP} and whole-body tracking~\cite{peng2018deepmimic}, which have contributed to improving anthropomorphic motion control capabilities.

\begin{figure}[]
    \centering
    \includegraphics[width=0.48\textwidth]{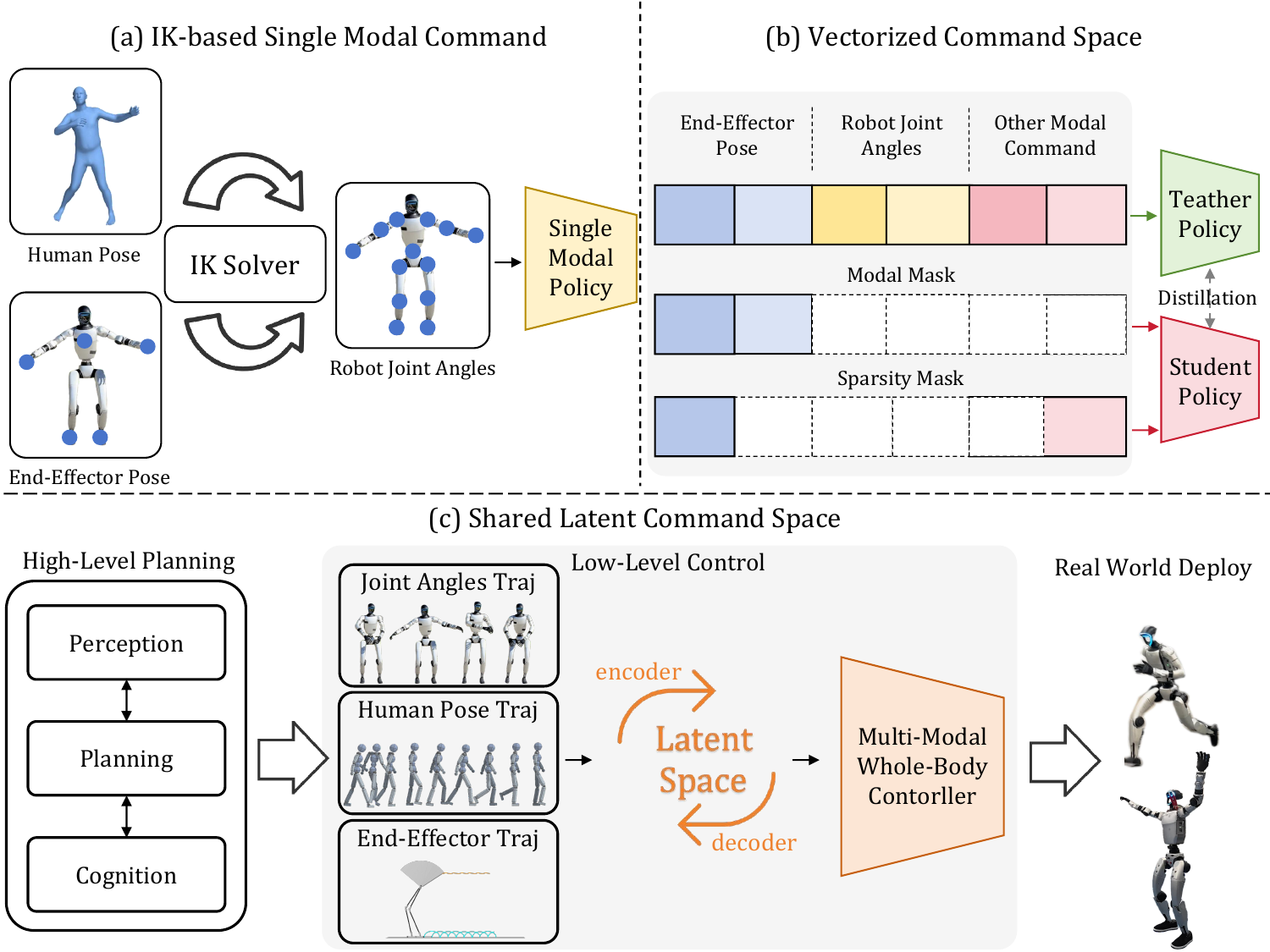}
    \caption{Without IK (a) or multi-stage distillation (b), we learn a shared latent command space and train a single policy end-to-end from heterogeneous motion references, enabling multi-modal whole-body control and real-world deployment.}
    \label{fig:cover}
\end{figure}

Building upon advances in motion stylization and tracking, human motion data~\cite{AMASS:ICCV:2019, li2023object, harvey2020robust, mason2022local} plays a central role in enabling humanoid robots to reproduce realistic behaviors.
These data are primarily collected through motion capture systems or virtual reality devices. The former provides dense, high-precision full-body joint trajectories, whereas the latter usually offers sparse end-effector trajectories better suited for interactive tasks.
\IEEEpubidadjcol
This sparsity inherently leads to kinematic underdetermination, where a single end-effector trajectory can correspond to multiple valid full-body configurations.
These differences reflect the multi-modal nature of human motion data in terms of representation structure and information granularity.
Leveraging heterogeneous motion modalities as references for humanoid control has been explored from different perspectives in prior works. Existing approaches generally follow two common strategies.
One representative line of work converts heterogeneous motion references into the target robot joint space using inverse kinematics (IK)~\cite{ze2025gmr}, thereby obtaining explicit robot joint angle trajectories that can be directly tracked or imitated~\cite{ze2025twist2, zhang2025track, han2025kungfubot2}.
Another line of work vectorizes and concatenates multi-modal motion references into a unified representation and trains a teacher policy with access to full-modality inputs. To accommodate modality absence during deployment, input masking is applied and a student policy is learned through teacher-student distillation, enabling operation under partial modality conditions~\cite{he2025hover}.
However, these two streams of works present certain limitations. The former requires additional inverse-kinematics computations during deployment for the conversion between different motion modalities. The latter concatenates vectorized multi-modal references without explicitly modeling cross-modal correlations and often requires a multi-stage training.

We aim to construct a versatile low-level humanoid controller through a unified multi-modal whole-body control framework, which supports multiple types of reference motions, including robot joint angles, human pose trajectories, and end-effector poses, as shown in Fig.~\ref{fig:cover}.
To achieve this goal, we face a primary challenge: learning a single unified policy from heterogeneous motion references that exhibit distinct sparsity patterns, disparate representation structures, and varying levels of training complexity.
Our method maps multi-modal motion references into a shared implicit command space for direct low-level control, rather than learning an implicit IK~\cite{chen2025implicit} replacement that collapses modality-specific sparsity patterns. And we introduce a curriculum learning strategy that progressively shifts from uniform sampling to failure-rate-based adaptive sampling, enabling balanced learning across motion references of different complexity.
Our key contributions are as follows:
\begin{itemize}
\item A unified multi-modal implicit command representation that enables a single end-to-end policy to directly consume heterogeneous motion references.
\item A multi-modal whole-body control framework with a curriculum-based sampling strategy that stabilizes training on large-scale motion datasets.
\item Comprehensive cross-modal generalization on unseen data, evidenced by a peak success rate of 98.42\% in simulation, enabling successful zero-shot sim-to-real deployment on a humanoid robot.
\end{itemize}

%% file: section/S2_relate_work.tex
\section{Related Work}

\subsection{Whole-Body Control for Humanoid Robots}

Physics-based simulation has driven significant advances in whole-body control for humanoid robots, with core challenges centering on sim-to-real transferability, challenging terrain~\cite{zhuang2024humanoid}, fall recovery~\cite{he2025learning}, the robustness of motion tracking~\cite{xie2025kungfubot}, the realization of highly dynamic motions~\cite{liao2025beyondmimic}, and highly balanced tasks~\cite{zhang2025hub}. 
Whole-body control methods for humanoid robots can be broadly categorized into two main approaches: task-driven and imitation-driven. Task-driven approaches primarily focus on enhancing autonomous locomotion capabilities in complex environments, particularly in uneven terrains~\cite{zhuang2024humanoid, he2025learning}. In contrast, imitation-driven methods aim to improve the robot motion performance and naturalness by mimicking human motion data. Within imitation-driven approaches, end-to-end motion stylization~\cite{2021-TOG-AMP} and motion tracking~\cite{2018-TOG-deepMimic} are two primary research directions.
Motion stylization methods use adversarial priors to encourage human-like behaviors while optimizing task rewards~\cite{2021-TOG-AMP,shi2025almi}. Motion tracking methods, originating from frameworks such as DeepMimic~\cite{2018-TOG-deepMimic}, directly imitate reference motion trajectories through pose and velocity matching rewards, providing a general foundation for learning diverse whole-body skills~\cite{he2025asap,xie2025kungfubot,liao2025beyondmimic}. Our work builds on this tracking-based paradigm and extends it to heterogeneous motion reference modalities.

\subsection{General Whole-Body Controller}

Humanoid robots require high-frequency feedback control to maintain balance and stability. Hierarchical control architectures have emerged as the dominant paradigm, decoupling low-frequency perception and planning at the high level from high-frequency stability control and motion execution at the low level. 
A key challenge is to build a universal low-level controller that can robustly support diverse high-level task specifications.
Hover~\cite{he2025hover} vectorizes and concatenates multi-modal motion commands, training an oracle policy using privileged information in the simulator. 
Through command masking and teacher-student distillation,
it trains a universal multi-modal motion tracking controller for diverse high-level tasks.
Most existing scalable controllers ultimately reduce heterogeneous motion guidance to joint-space targets.
For example, Any2Track~\cite{zhang2025track} employs clustering to categorize data into multiple motion classes, trains separate sub-policies, and subsequently distills them into a single multi-motion tracking policy.
KungfuBot2~\cite{han2025kungfubot2} uses Orthogonal Mixture-of-Experts (OMoE) and segment-level rewards to disentangle skills and enable efficient multi-skill learning.
TWIST2~\cite{ze2025twist2} converts observations from inertial motion capture devices into dense reference inputs, enabling robust teleoperation in unseen scenarios.
Sonic~\cite{luo2025sonic} introduces a tokenized unified motion
representation for multi-modal references and demonstrates the scaling
potential of humanoid motion tracking on large-scale motion data.
In contrast, we treat multi-modal motion references as different
observations of the same underlying motion intent and explicitly learn a
continuous shared latent command space, enabling us to align
heterogeneous references while analyzing how reference modality affects
tracking fidelity and robustness in downstream settings such as motion
tracking, teleoperation, and loco-manipulation.

%% file: section/S3_method.tex
\section{Method}

\begin{figure*}[htbp]
    \centering
    \includegraphics[width=\textwidth]{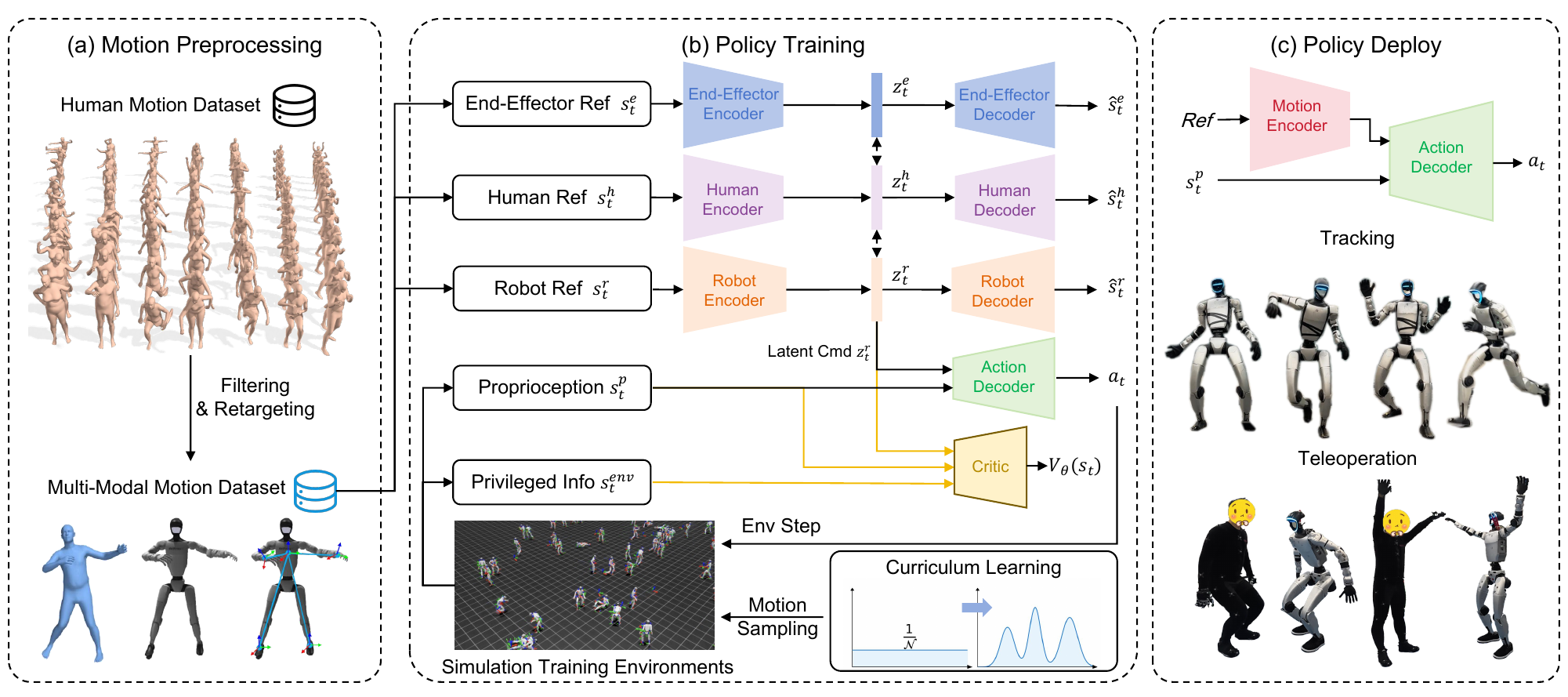}
    \caption{Overview of the \oursmethod~framework. 
    (a) We filter and preprocess large-scale human motion datasets into multi-modal reference.
    (b) The multi-modal command encoder processes different types of reference into a unified latent space. 
    (c) The policy is trained once and deployed across multiple input modalities.
    }
    \label{fig:method_overview}
\end{figure*}

In this section, we detail the problem formulation, multi-modal policy learning framework, curriculum-based sampling strategy, and sim-to-real transfer pipeline. The overall training and deployment framework is illustrated in Fig.~\ref{fig:method_overview}.

\subsection{Problem Definition}

We formulate whole-body contorl as a reinforcement learning problem modeled as a Markov Decision Process (MDP) $\mathbf{M} = (\mathbf{S}, \mathbf{A}, P, r, \gamma)$.
At each time step $t$, the state $\mathbf{s}_t \in \mathbf{S}$ consists of the robot proprioceptive state $\mathbf{s}_t^p$ and a reference motion state $\mathbf{s}_t^g$ extracted from a target motion sequence, with $\mathbf{s}_t = (\mathbf{s}_t^p, \mathbf{s}_t^g)$.
Given $\mathbf{s}_t$, the policy $\pi(\mathbf{a}_t \mid \mathbf{s}_t)$ outputs an action $\mathbf{a}_t \in \mathbf{A}$. 
The environment transitions to $\mathbf{s}_{t+1} \sim P(\cdot \mid \mathbf{s}_t, \mathbf{a}_t)$ and produces a reward $r_t = r(\mathbf{s}_t, \mathbf{a}_t)$, which encourages accurate tracking of the reference motion while maintaining physical feasibility.
The objective is to learn a policy that maximizes the expected discounted return:
\begin{equation}
    \max_{\pi} 
    \mathbb{E}_{\xi \sim \pi} 
    \left[ 
        \sum_{t=0}^{T-1}
        \gamma^t r(\mathbf{s}_t, \mathbf{a}_t) 
    \right],
\end{equation}
where $\xi = (\mathbf{s}_0, \mathbf{a}_0, \dots, \mathbf{s}_{T-1}, \mathbf{a}_{T-1})$ denotes a trajectory generated by policy $\pi$, and $T$ is the episode horizon.

\subsubsection{Multi-Modal Reference Encoding}

We consider three motion modalities to support diverse data sources and tasks: robot joint angles $\mathbf{q}_t$, human body pose $\boldsymbol{\phi}_t^{\text{smplx}}$, and end-effector poses $\mathbf{T}_t$. Here, $\mathbf{q}_t \in \mathbb{R}^{29}$ denotes the 29-dimensional robot joint angles for precise joint-level tracking. $\boldsymbol{\phi}_t^{\text{smplx}} \in \mathbb{R}^{21 \times 6}$ represents the SMPL-X body pose~\cite{SMPL-X:2019}, where each of the 21 joints is converted from axis-angle to a 6D rotation representation to leverage large-scale human datasets. $\mathbf{T}_t \in \mathbb{R}^{5 \times 9}$ encodes the $SE(3)$ poses of $M=5$ end-effectors (feet, hands, and chest), including 3D root-relative positions and 6D rotation representations for sparse teleoperation scenarios. At each time step $t$, the autoencoder takes as input a short-horizon reference sequence for each modality.
Specifically, we define the reference inputs $\mathbf{s}_t^g$ as
$
\mathbf{s}_t^r = \{\mathbf{q}_\tau\}_{\tau \in h_t},
\mathbf{s}_t^h = \{\boldsymbol{\phi}_\tau^{\text{smplx}}\}_{\tau \in h_t},
\mathbf{s}_t^e = \{\mathbf{T}_\tau\}_{\tau \in h_t}.
$
The temporal horizon is defined as
$
h_t := \{\, t + k\Delta \mid k = 0, \dots, H-1 \,\},
$
where $\Delta$ denotes the sampling interval and $H$ represents the horizon length.
We set $H=10$ and $\Delta=2$ to balance responsiveness and stability,
allowing the encoder to capture short-term motion dynamics while maintaining control smoothness.
For each modality, a dedicated encoder extracts a latent representation
$\mathbf{z}_t^m \in \mathbb{R}^{64}$, where $m\in\{r,h,e\}$
which is subsequently used for reconstruction.

\subsubsection{Asymmetric Actor-Critic}

To improve policy learning efficiency, we adopt an asymmetric actor-critic architecture. Specifically, the actor network utilizes only partial observation information, which we can obtain during real-world deployment, to select actions. The actor input state $\mathbf{s}_t=(\mathbf{z}_t, \mathbf{s}^p_t)$, where $\mathbf{s}^p_t=(\Delta \boldsymbol{\theta}_t, \boldsymbol{\omega}_t, \mathbf{q}_t, \dot{\mathbf{q}}_t, \mathbf{a}_{t-1})$ represents the robot's proprioceptive state including root rotation error $\Delta \boldsymbol{\theta}_t \in \mathbb{R}^{6}$ to maintain heading alignment.
Notably, we deliberately exclude global root position and linear velocity from $\mathbf{s}_t$ to ensure robustness in non-instrumented environments without external localization.
The critic network has access to more comprehensive environment information to provide privileged supervision; the critic input state $\mathbf{s}_t^{\text{critic}}=(\mathbf{s}_t, \mathbf{s}_t^{\text{env}})$ includes the full environment state $\mathbf{s}_t^{\text{env}}=(\Delta \mathbf{p}_t, \mathbf{\mathcal{B}}_p, \mathbf{\mathcal{B}}_\theta, \mathbf{v}_t )$, where $\Delta \mathbf{p}_t\in \mathbb{R}^{3}$ is the root position error; $\mathbf{\mathcal{B}}_p\in \mathbb{R}^{42}, \mathbf{\mathcal{B}}_\theta\in \mathbb{R}^{84}$ are the body link positions and orientations; and $\mathbf{v}_t\in \mathbb{R}^{3}$ represents root linear velocity. The critic network outputs a scalar value $\mathbf{V}_{\boldsymbol{\theta}}(\mathbf{s}_t) \in \mathbb{R}$.

\subsubsection{Policy Training}

At each time step $t$, the reference sequences from different modalities, robot joint angles $\mathbf{s}_t^r$, human body pose $\mathbf{s}_t^h$, and end-effector poses $\mathbf{s}_t^e$, 
are encoded by modality-specific encoders
$\mathcal{E}_r, \mathcal{E}_h, \mathcal{E}_e$
into latent representations
$\mathbf{z}_t^r, \mathbf{z}_t^h, \mathbf{z}_t^e$.
Each latent code is decoded by the corresponding decoder
$\mathcal{D}_r, \mathcal{D}_h, \mathcal{D}_e$
to reconstruct $\hat{\mathbf{s}}_t^r, \hat{\mathbf{s}}_t^h, \hat{\mathbf{s}}_t^e$.
All encoders and decoders are implemented as multi-layer perceptrons (MLPs).
The training objective combines reinforcement learning and multi-modal representation learning:
\begin{equation}
\mathcal{L} = \mathcal{L}_\pi + \mathcal{L}_{\mathrm{ae}},
\end{equation}
where $\mathcal{L}_\pi$ is the policy loss and $\mathcal{L}_{\mathrm{ae}}$ is the multi-modal autoencoder loss.

\textit{Policy Loss:} 
We adopt a clipped surrogate objective with value function and entropy regularization:
\begin{align}
\mathcal{L}_\pi
=& \mathbb{E}_t \Big[
-\min \Big( r_t(\boldsymbol{\theta}) \hat{A}_t,
\operatorname{clip}(r_t(\boldsymbol{\theta}), 1-\epsilon, 1+\epsilon)
\hat{A}_t \Big) \notag \\
& + \lambda_c
\|\mathbf{V}_{\boldsymbol{\theta}}(\mathbf{s}_t)
- \mathbf{V}_t^\mathrm{target}\|_2^2
 - \lambda_h \,
\mathcal{H}(\pi_{\boldsymbol{\theta}}(\cdot \mid \mathbf{s}_t))
\Big].
\end{align}
where $\mathbf{s}_t$ denotes the actor input state, 
$\hat{A}_t$ is the advantage estimate, 
$\mathbf{V}_{\boldsymbol{\theta}}$ is the value function, 
$\lambda_c$ is the value loss coefficient that scales the squared error between the predicted and target values, 
$\lambda_h$ is the entropy regularization coefficient that encourages exploration, 
and $\mathcal{H}(\cdot)$ denotes the policy entropy~\cite{sutton1998reinforcement}.

\textit{Multi-Modal Autoencoder Loss:} 
To ensure consistent latent representations across modalities, we define:

\begin{equation}
\begin{aligned}
\mathcal{L}_{\mathrm{ae}} &= \lambda_\mathrm{recon} \mathcal{L}_\mathrm{recon}
+ \lambda_\mathrm{align} \mathcal{L}_\mathrm{align}
+ \lambda_\mathrm{cons} \mathcal{L}_\mathrm{cons}, \\
\mathcal{L}_\mathrm{recon} &= \sum_{m \in \{r,h,e\}} \|\mathbf{s}_t^m - \hat{\mathbf{s}}_t^m\|_2^2, \\
\mathcal{L}_\mathrm{align} &= \sum_{(m,n) \in \{(r,h),(r,e),(h,e)\}} \|\mathbf{z}_t^m - \mathbf{z}_t^n\|_2^2, \\
\mathcal{L}_\mathrm{cons} &= \sum_{(m,n) \in \{(r,h),(r,e),(h,e)\}} \|\mathcal{D}_r(\mathbf{z}_t^m) - \mathcal{D}_r(\mathbf{z}_t^n)\|_2^2.
\end{aligned}
\end{equation}
Here, $\mathcal{L}_\mathrm{recon}$ enforces accurate reconstruction for each modality, 
$\mathcal{L}_\mathrm{align}$ encourages latent representations from different modalities to be close, 
and $\mathcal{L}_\mathrm{cons}$ ensures decoded outputs are consistent when using a shared decoder. 
The hyperparameters $\lambda_\mathrm{recon}, \lambda_\mathrm{align}, \lambda_\mathrm{cons}$ balance the contributions of each term.

The reward and regularization terms used for training are summarized in Table~\ref{tab:reward_terms}. 

\input{table/table_reward.tex}

\subsection{Curriculum Learning Strategy}

Learning from large-scale motion data is challenging due to heterogeneous data distribution and varying intrinsic motion complexity. We adopt a curriculum learning strategy that adaptively reshapes the sampling distribution during training.
Each trajectory is divided into 1-second segments. During training, we maintain an exponential moving average (EMA) of the termination failure rate for each segment, which serves as a measure of its execution difficulty. Sampling probabilities are computed accordingly, progressively shifting focus toward segments with higher failure rates.
To stabilize early-stage training, we blend failure-based sampling with uniform sampling using a mixing coefficient $\alpha$, which is gradually decreased over time. Additionally, we clip the probabilities of segments with extremely high failure rates to avoid over-emphasizing physically infeasible motions.
The final sampling probability is defined as
\begin{equation}
    P_{\text{final}} = \alpha P_{\text{uniform}} + (1-\alpha) \text{clip}(P_{\text{failure}}, 0, P_{\text{max}}),
\end{equation}
where $\text{clip}(\cdot)$ limits the sampling probability within $[0, P_{\text{max}}]$ to ensure stability.

\subsection{Sim-to-Real Transfer}

To facilitate sim-to-real transfer, we employ domain randomization during training. We randomly vary physical parameters such as ground friction coefficients, link masses and inertias, joint damping and motor strength, and we inject observation noise into joint encoder and inertial sensor readings. 
These randomizations expose the policy to a wide range of dynamics and sensing conditions, thereby improving its robustness to modeling errors, unmodeled contacts, and hardware variability during real-world deployment.
The applied domain randomization settings are summarized in Table \ref{tab:domain_randomization}.

\input{table/table_domain_randomization.tex}

%% file: table/table_reward.tex
\begin{table}[htbp]
\centering
\caption{Reward terms used in RL training.}
\label{tab:reward_terms}
\begin{tabular}{lcc}
\toprule
\textbf{Term}                  & \textbf{Expression}                                                                                                                              & \textbf{Weight} \\
\midrule
Root velocity         & $\exp\left( -\frac{\left\| \mathbf{v}^{\text{ref\_root}} - \mathbf{v}^{\text{root}} \right\|^2}{\sigma_{\text{root\_vel}}^2} \right)$                     & 0.5    \\
Root orientation      & $\exp\left( -\frac{\left\| \mathbf{\theta }^{\text{ref\_root}} - \mathbf{\theta }^{\text{root}} \right\|^2}{\sigma_{\text{root\_rot}}^2} \right)$         & 0.5    \\
Body position         & $\exp\left( -\frac{\sum_{i=1}^{N} \left\| \mathbf{p}^{\text{ref}}_i - \mathbf{p}_i \right\|^2}{N\cdot\sigma_{\text{b\_pos}}^2} \right)$                   & 1.0    \\
Body orientation      & $\exp\left( -\frac{\sum_{i=1}^{N} \left\| \mathbf{\theta }^{\text{ref}}_i - \mathbf{\theta}_i \right\|^2}{N\cdot\sigma_{\text{b\_rot}}^2} \right)$        & 1.0    \\
Body velocity         & $\exp\left( -\frac{\sum_{i=1}^{N} \left\| \mathbf{v}^{\text{ref}}_i - \mathbf{v}_i \right\|^2 }{N\cdot\sigma_{\text{b\_vel}}^2}\right)$                   & 1.0    \\
Body angular velocity & $\exp\left( -\frac{\sum_{i=1}^{N} \left\| \boldsymbol{\omega}^{\text{ref}}_i - \boldsymbol{\omega}_i \right\|^2}{N\cdot\sigma_{\text{b\_ang}}^2}\right)$  & 1.0    \\
\midrule
Action rate           & $||a_t-a_{t-1}||_2^2$                                                                                                                                     & -0.1   \\
Joint pos limits      & $\mathbb{I}(q \notin [q_{\text{soft-min}}, q_{\text{soft-max}}])$                                                                                         & -10.0  \\
Collision             & $\mathbb{I}_{\text{collision}}$                                                                                                                           & -0.1   \\
\bottomrule
\end{tabular}
\end{table}

%% file: table/table_domain_randomization.tex
\begin{table}[htbp]
\centering
\caption{Domain randomization and noised observation terms. $^*$ indicates the maximum push magnitude across all directions.}
\label{tab:domain_randomization}
\resizebox{0.48\textwidth}{!}{
\begin{tabular}{cccc}
\toprule
DR terms & Range & Noised Obs & Range \\
\midrule
Friction (-) & {[}0.1, 1.6{]} & Anchor ori (rad) & {[}-0.05, 0.05{]} \\
Push robot$^*$ (m/s) & {[}-0.5, 0.5{]}  & Base ang vel (rad/s) & {[}-0.2, 0.2{]} \\
Base COM (m) & {[}-0.1, 0.1{]} & Joint pos (rad) & {[}-0.01, 0.01{]} \\
Base mass (kg) & {[}-0.8, 1.2{]} & Joint vel (rad/s) & {[}-0.5, 0.5{]} \\
Default joint pos (rad) & {[}-0.01, 0.01{]} & & \\
\bottomrule
\end{tabular}}
\end{table}

%% file: section/S4_experiments.tex
\section{Experiments}

In this section, we present experimental results in both the simulator and the real-world deployment. Our experimental platform is the 29-DoF Unitree G1 humanoid robot.
Here, we aim to answer the following research questions:
\begin{itemize}
    \item \textbf{Q1:} Does the choice of input modality significantly affect deploy performance?
    \item \textbf{Q2:} How do model size and dataset scale influence large-scale training?
    \item \textbf{Q3:} How robustly does the autoencoder generalize to out-of-distribution (OOD) motion sequences?
\end{itemize}

\subsection{Experiments Setup}

We use IsaacSim~\cite{NVIDIA_Isaac_Sim} as the simulator to train our model. We use Blender software to refine the SMPL-X human body model and employ GMR~\cite{ze2025gmr} for data retargeting.
We optimized the model with the RAD~\cite{lyu2025conformal} optimizer.

\textbf{Datasets:}
We select LAFAN1~\cite{harvey2020robust}, which contains diverse and highly dynamic human motions, and 100STYLE~\cite{mason2022local}, which features a wide range of walking styles, as our training datasets. The test set is the test subset of the OMOMO~\cite{li2023object} dataset, which features complex manipulation.

\textbf{Baselines:} We select HOVER~\cite{he2025hover}, ExBody2~\cite{ji2024exbody2}, OmniH2O~\cite{he2024omnih2o} and TWIST2~\cite{ze2025twist2} as baseline methods for comparison.
Hover is a multi-modal command tracking framework based on command masking and teacher-student distillation.
ExBody2 is a general tracking control framework based on multi-expert distillation.
OmniH2O is a general humanoid whole-body motion tracking framework that employs a teacher-student learning paradigm.
TWIST2 is a teleoperation oriented framework that retargets human observations into robot references.
To ensure authenticity, all baselines were evaluated using their original, unmodified Domain Randomization (DR) settings.

\textbf{Metrics:} We evaluate tracking accuracy using root-relative mean per-keypoint position error ($E_{\text{mpkpe}}$, mm), root-relative mean per-joint angle error ($E_{\text{mpjae}}$, rad), mean root linear velocity error ($E_{\text{vel}}$, m/s),
and episode success rate (\%).
Starting poses are randomly sampled from the trajectories. An episode is considered successful if the agent maintains its root orientation within ±45 degrees of the reference throughout the entire episode.

\textbf{Deployment:}
We evaluate the trained policies under three input modalities: 
robot joint angles, human body pose, and end-effector poses. 
The corresponding deployed policies are denoted as 
$\pi^r$, $\pi^h$, and $\pi^e$, 
respectively. 
Unless otherwise specified, the results reported use the robot reference policy $\pi^r$. 
During simulation testing, we use the same environment configuration as during training.

\subsection{Main Results}

We first evaluate our method on the training dataset and compare it with the baselines.
During evaluation, we disable the adaptive sampling strategy used in training to ensure a fair and consistent comparison across all methods.
All metrics are computed from 10,000 simulation steps collected over 4096 parallel environments.
For HOVER, we report the validation results of the teacher policy.

As reported in Table~\ref{tab:main_metrics}, M3imic achieves the best
overall performance on the training datasets. It reaches a success rate
of 99.54\%, slightly higher than TWIST2 (98.89\%) and higher than
ExBody2 (98.12\%) and OmniH2O (97.78\%). In terms of tracking accuracy,
M3imic obtains $E_{\text{mpkpe}}=46.05$~mm,
$E_{\text{mpjae}}=0.112$~rad, and $E_{\text{vel}}=0.256$~m/s.
Compared with the strongest baseline TWIST2, this corresponds to
reductions of 10.8\%, 7.4\%, and 4.1\% in $E_{\text{mpkpe}}$,
$E_{\text{mpjae}}$, and $E_{\text{vel}}$, respectively.
Compared with ExBody2 and OmniH2O, the reduction in
$E_{\text{mpkpe}}$ is 13.5\% and 26.6\%, respectively.

\input{table/table_train_metrics.tex}

We further evaluate the three deployed policies on the unseen OMOMO test
dataset, as summarized in Table~\ref{tab:test_metrics}. The robot-joint
policy $\pi^r$ achieves the best pose and joint-angle tracking accuracy
($E_{\text{mpkpe}}=71.52$~mm, $E_{\text{mpjae}}=0.139$~rad), reducing
these errors by 8.8\%/2.8\% over TWIST2 and 13.3\%/4.1\% over ExBody2.
In contrast, the end-effector policy $\pi^e$ achieves the highest success
rate of 98.42\%, outperforming TWIST2, ExBody2, OmniH2O, and $\pi^r$ by
3.83, 3.74, 5.58, and 2.44 percentage points, respectively. This shows
that dense joint references favor tracking fidelity, while sparse
end-effector references provide better robustness under distribution
shift. The human-pose policy $\pi^h$ performs comparably to $\pi^r$.

\begin{figure}[!h]
    \centering
    \includegraphics[width=0.48\textwidth]{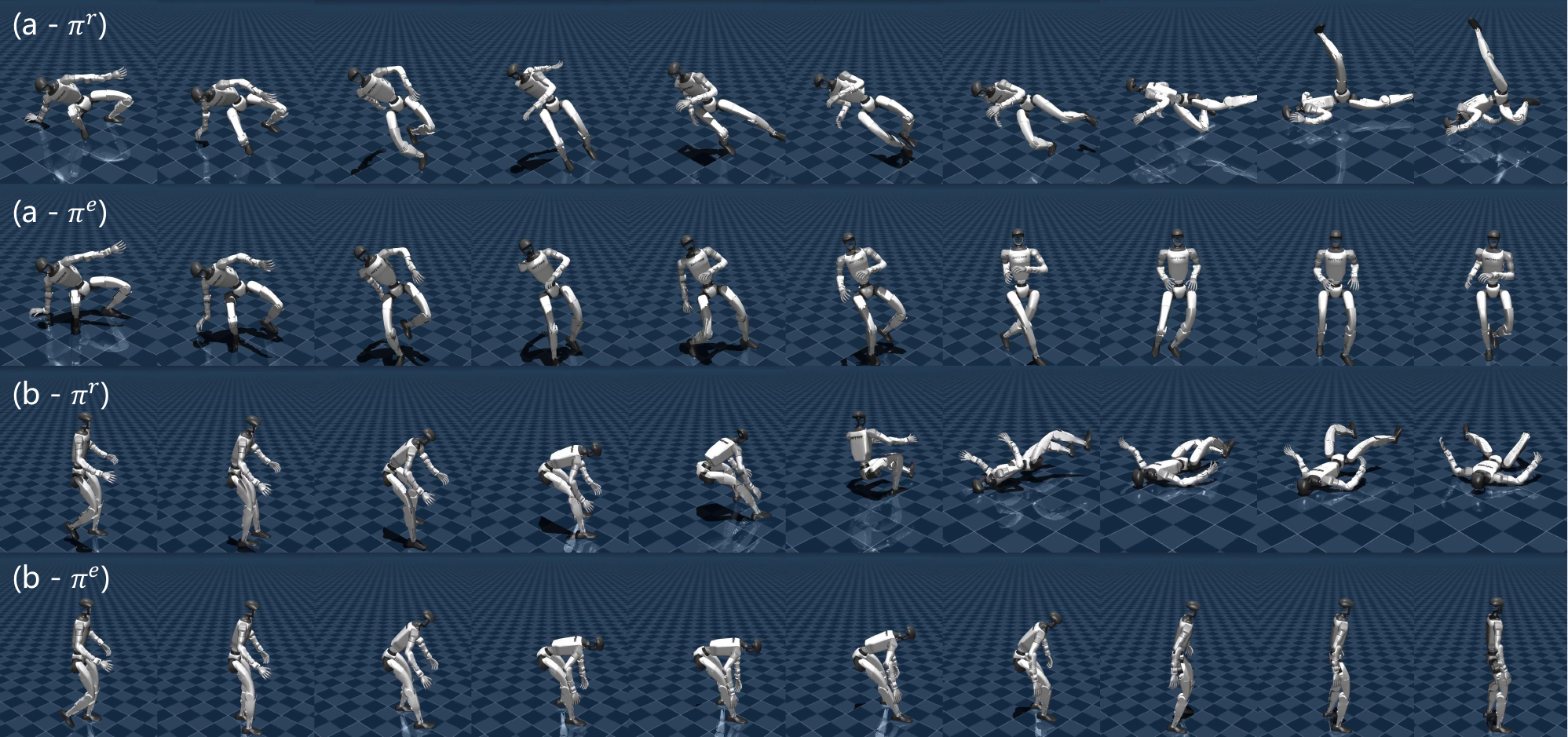}
    \caption{Qualitative comparison between $\pi^r$ and $\pi^e$ in MuJoCo simulation.}
    \label{fig:sim2sim_case}
\end{figure}

We further analyze the modality differences in Table~\ref{tab:test_metrics}
and provide a visual comparison in MuJoCo in Fig.~\ref{fig:sim2sim_case}.
The dense constraints of $\pi^r$ favor local pose accuracy but leave less
freedom for balance adjustment, while $\pi^e$ exploits the kinematic
redundancy of sparse end-effector references to achieve higher robustness.
This reveals a fidelity-robustness trade-off among modalities, thereby
answering Q1.

\input{table/table_test_metrics.tex}

These results suggest that the improvement comes from three complementary
designs: single-stage training avoids the information loss, distribution
mismatch, and accumulated errors introduced by policy transfer in
multi-stage distillation; failure-rate-based adaptive sampling improves
training efficiency by emphasizing difficult motion segments; and the
multi-modal design balances tracking fidelity from dense joint references
with robustness from sparse end-effector references under distribution
shift.

\begin{figure*}[ht]
    \centering
    \includegraphics[width=\textwidth]{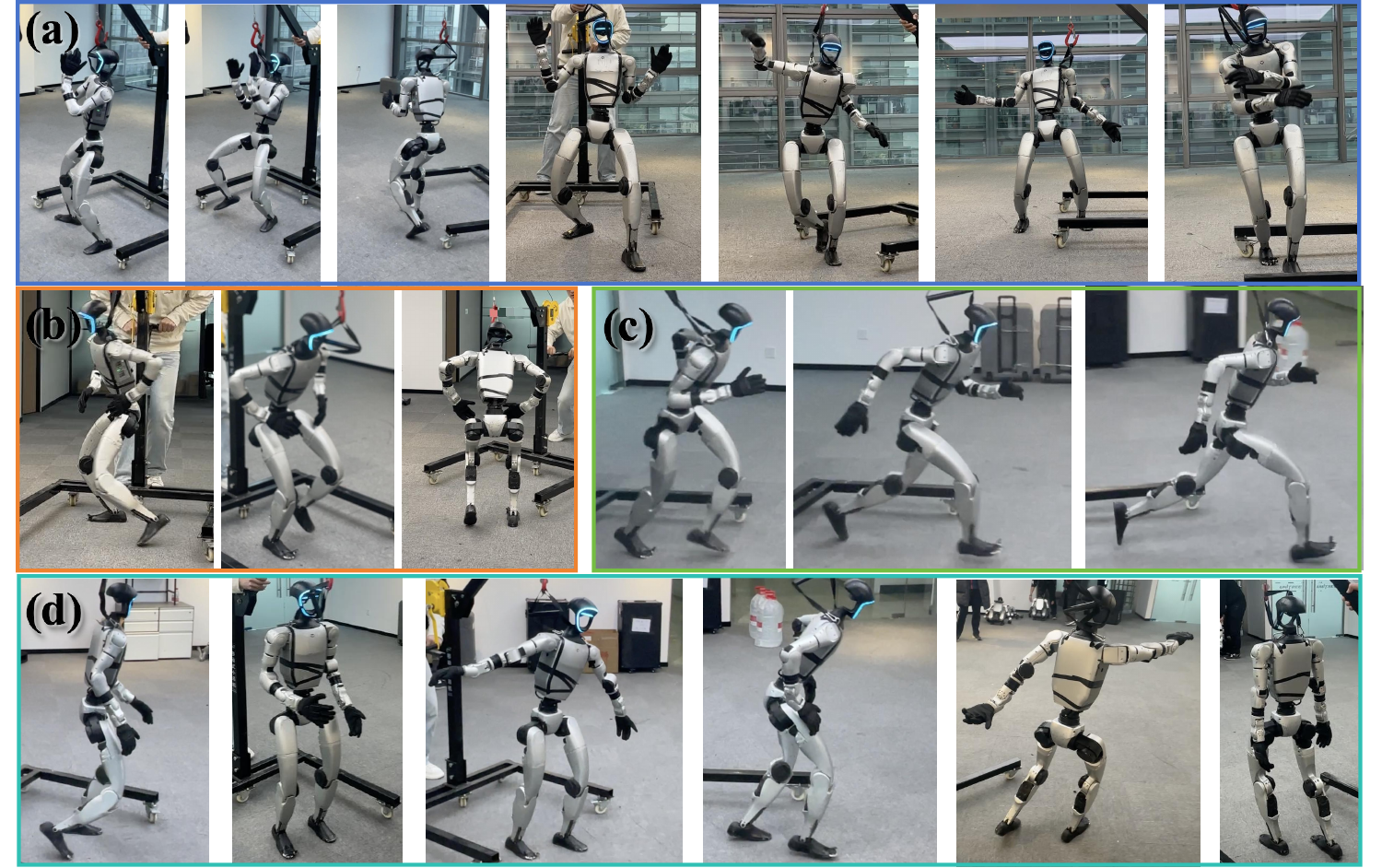}
    \caption{
    We use the same policy network for diverse action tracking in real-world environment.
    (a) The performance of humanoid robot in high-dynamic dance motions.
    (b) The humanoid robot performs backward walking. 
    (c) The humanoid robot is running at high speed.
    (d) The humanoid robot tracks a wide range of walking postures.}
    \label{fig:exp_tracking}
\end{figure*}

\begin{figure}[ht]
    \centering
    \includegraphics[width=0.44\textwidth]{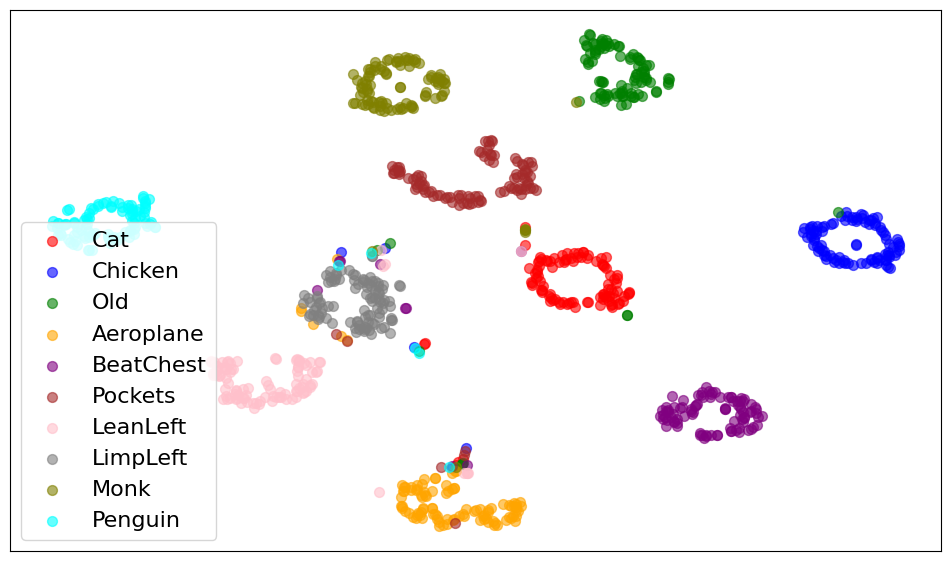}
    \caption{t-SNE visualization of the latent space distribution for different motions categories. The labels ``Cat", ``Chicken", etc. indicate the corresponding motion categories.}
    \label{fig:data_t-sne}
\end{figure}

\begin{figure}[ht]
    \centering
    \includegraphics[width=0.44\textwidth]{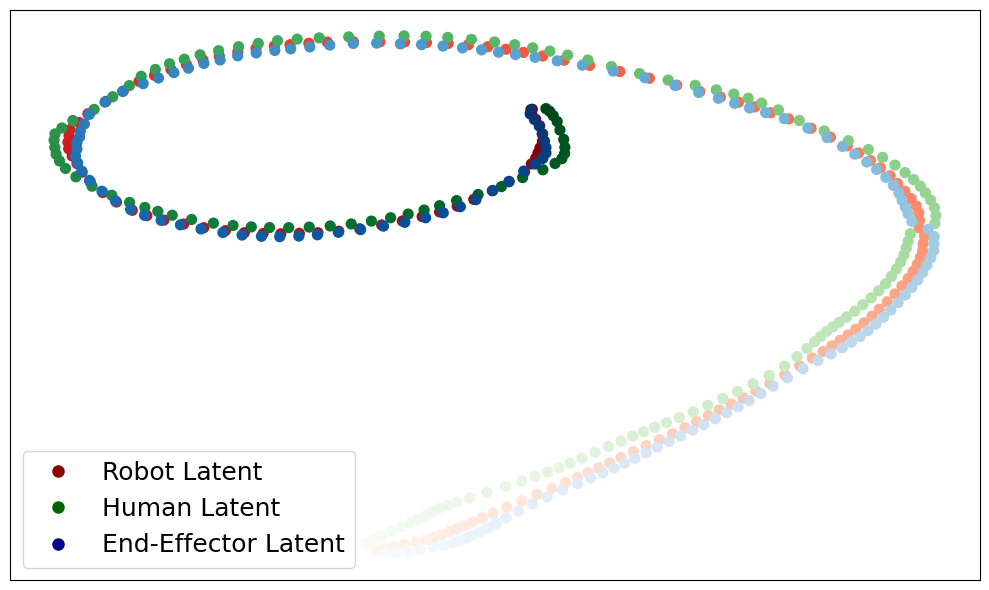}
    \caption{t-SNE visualization of the latent space distribution for different input modalities of the same motion. The color variation represents time. }
    \label{fig:multi-modal_t-sne}
\end{figure}

We visualize the latent space distribution after multi-modal encoding using t-SNE~\cite{maaten2008visualizing}. 
Fig.~\ref{fig:data_t-sne} illustrates the latent space distribution of motions in the 100STYLE dataset, demonstrating that the multi-encoder architecture effectively separates distinct motion categories.
Fig.~\ref{fig:multi-modal_t-sne} shows the latent space distribution for different input modalities of the same motion, where the distributions for different modalities exhibit clear consistency, indicating that the multi-modal encoder architecture effectively integrates information from different input modalities.

\begin{figure*}[!ht]
    \centering
    \includegraphics[width=\textwidth]{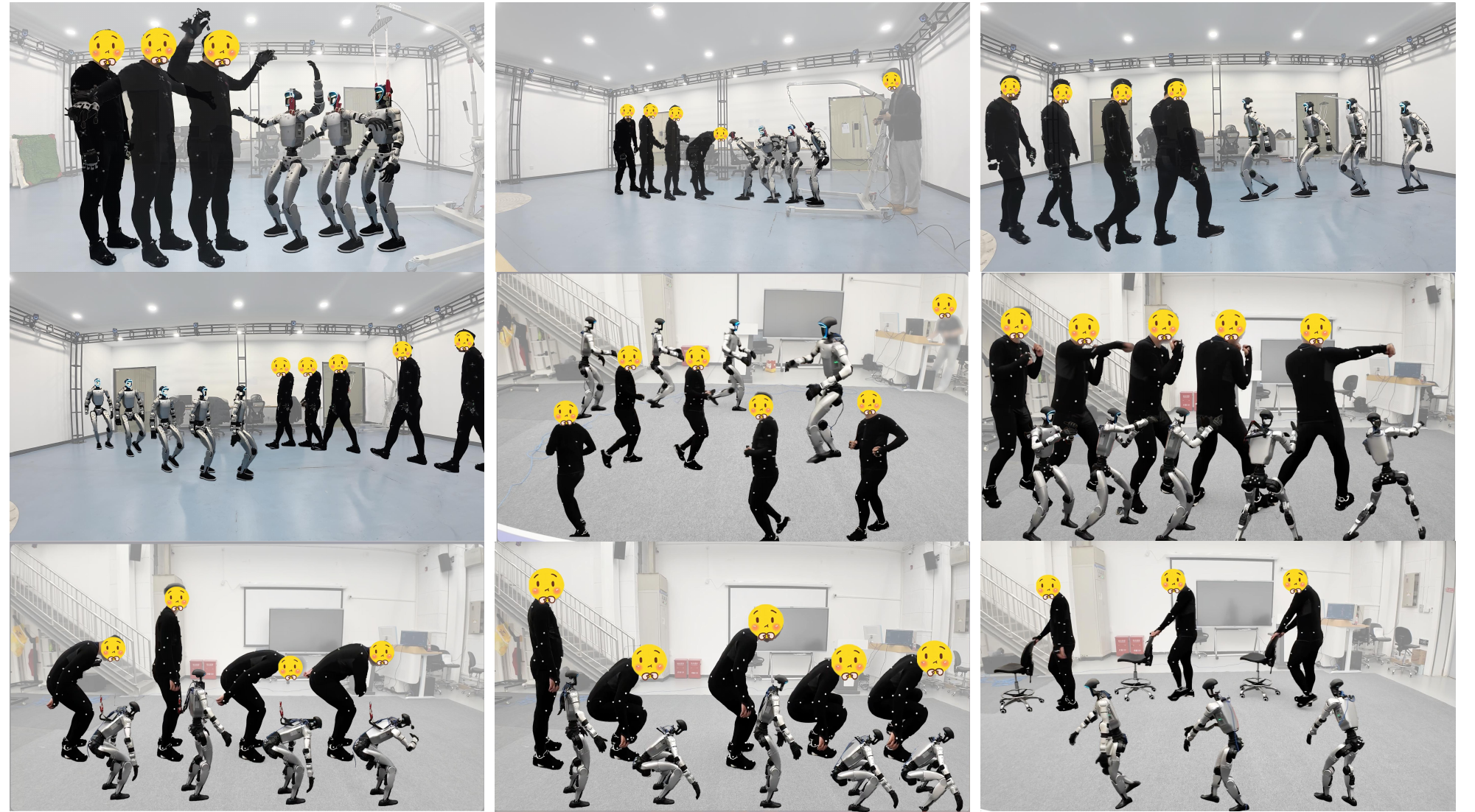}
    \caption{Real-world teleoperation experiments using an optical motion capture system. 
    We demonstrated a variety of motions including bending over, raising hands, walking, jogging, boxing, squatting, and pushing objects.}
    \label{fig:exp_teleoperation}
\end{figure*}

\subsection{Ablation Study}

To analyze the contribution of key design choices, we conduct ablation studies on curriculum learning, dataset size, and network size using the 100STYLE dataset. All ablation models are trained for 50,000 iterations on 4 RTX 4090 GPUs. The results are reported in Table.~\ref{tab:ablation} and Table.~\ref{tab:ablation_dataset}.

\input{table/table_ablation.tex}

\input{table/table_ablation_network.tex}

\subsubsection{Ablation on Curriculum Learning Strategy}

For curriculum learning, we use $\alpha_{min}=0.2$ and $P_{\text{max}}=0.05$. The ablated setting fixes the uniform sampling ratio to $\alpha=1.0$, corresponding to training without curriculum learning. As shown in Table.~\ref{tab:ablation}, curriculum learning consistently improves the task success rate while maintaining comparable tracking accuracy. This suggests that adaptive sampling mainly benefits training robustness rather than simply optimizing short-term tracking errors.

\subsubsection{Ablation on Dataset Size}

We compare models trained on the full dataset, containing approximately 3.9 million frames, with models trained on 10\% of the data.
Curriculum learning is disabled in this comparison to isolate the effect of data scale. As shown in Table.~\ref{tab:ablation_dataset}, increasing the dataset size improves all tracking and success metrics on the OMOMO test set, indicating that data scale is critical for generalization to unseen motions.

\subsubsection{Ablation on Network Size}

The standard model uses MLP layers of sizes [512, 256, 128], while the large model uses [1024, 512, 256]. Increasing network capacity improves performance, especially in the low-data setting, but the gains become smaller when the full dataset is used. Meanwhile, the larger network reduces training and inference throughput, as reflected by the lower relative speed in Table.~\ref{tab:ablation}. These results indicate that dataset scale has a stronger impact than network size, while larger models introduce a clear efficiency trade-off, thereby answering Q2.

\subsection{Real-World Deployment}

To answer Q3, we deployed \oursmethod~ in a real-world environment without collecting any teleoperation data or task-specific datasets for training. We clarify that OOD shifts may come from either input-domain differences, such as sensor noise and retargeting-induced pose bias, or unseen motion categories. In this work, we mainly focus on the former, evaluating robustness to real-world teleoperation commands that differ from the retargeted references used during training.

\subsubsection{In-domain Tracking}

As shown in the Fig.~\ref{fig:exp_tracking}, our method can effectively track the motions from the training dataset in a real-world environment, demonstrating strong robustness.
For the deployment process, we utilized the robot encoder as the motion encoder. We demonstrated tracking performance across various high-dynamic actions, including dancing, running, and walking.
These experiments showcase the strong robustness of our method in real-world environments.

\subsubsection{Out-of-domain Teleoperation}

For teleoperation tasks, we employed an optical motion capture system to obtain end-effector data from human operators. 
The Fig.~\ref{fig:exp_teleoperation} illustrates our teleoperation tasks in a real-world environment, showcasing the generalization capabilities of our model.
We compare two deployment policies, $\pi^r$ and $\pi^e$, using motions
collected from the same source. The collected motions are converted into
dense robot-joint references for $\pi^r$ and sparse end-effector
references for $\pi^e$, respectively. The average real-world tracking
results are reported in Table~\ref{tab:exp_real_avg}. $\pi^r$ achieves
slightly lower tracking errors, with $E_{\text{mpkpe}}=41.63$~mm,
$E_{\text{mpjae}}=0.095$~rad, and $E_{\text{vel}}=0.260$~m/s, while
$\pi^e$ achieves comparable performance using only sparse end-effector
commands. These results indicate that $\pi^e$ provides a practical
interface for teleoperation while maintaining reasonable whole-body
tracking accuracy under real-world distribution shifts.

\begin{table}[!ht]
\centering
\caption{Average real-world evaluation results of $\pi^r$ and $\pi^e$.}
\label{tab:exp_real_avg}
\begin{tabular}{c|ccc}
\toprule
Policy & $E_{\text{mpkpe}}$ (mm)$\downarrow$ 
       & $E_{\text{mpjae}}$ (rad)$\downarrow$ 
       & $E_{\text{vel}}$ (m/s)$\downarrow$ \\
\midrule
$\pi^r$ & 41.63 & 0.095 & 0.260 \\
$\pi^e$ & 43.22 & 0.105 & 0.268 \\
\bottomrule
\end{tabular}
\end{table}

According to the experimental results, we recommend enhancing model generalization by increasing the diversity of motions rather than merely the quantity, especially by incorporating datasets from different sources, thereby affirmatively answering Q3.

%% file: table/table_train_metrics.tex
\begin{table}[!ht]
\centering
\caption{Main Metrics in Training Dataset LAFAN1~\cite{harvey2020robust} and 100STYLE~\cite{mason2022local}.}
\label{tab:main_metrics}
\resizebox{0.48\textwidth}{!}{
\begin{tabular}{ccccc}
\toprule
Method  & Success (\%) $\uparrow$ & $E_{\text{mpkpe}}$ (mm) $\downarrow$ & $E_{\text{mpjae}}$ (rad) $\downarrow$ & $E_{\text{vel}}$ (m/s) $\downarrow$ \\
\midrule
HOVER   & 87.35$_{\pm 0.26}$ & 128.20$_{\pm 3.54}$ & 0.686$_{\pm 0.025}$ & 0.481$_{\pm 0.044}$  \\
ExBody2 & 98.12$_{\pm 0.09}$ & 53.25$_{\pm 2.52}$  & 0.146$_{\pm 0.016}$ & 0.285$_{\pm 0.023}$ \\
OmniH2O & 97.78$_{\pm 0.10}$ & 62.75$_{\pm 3.23}$. & 0.154$_{\pm 0.010}$ & 0.307$_{\pm 0.025}$ \\
TWIST2  & 98.89$_{\pm 0.08}$ & 51.65$_{\pm 1.84}$  & 0.121$_{\pm 0.009}$ & 0.267$_{\pm 0.015}$ \\
Ours    & \textbf{99.54$_{\pm 0.04}$} &\textbf{46.05$_{\pm 2.07}$} & \textbf{0.112$_{\pm 0.013}$} & \textbf{0.256$_{\pm 0.019}$} \\
\bottomrule
\end{tabular}
}
\end{table}

%% file: table/table_test_metrics.tex
\begin{table}[!ht]
\centering
\caption{Main Metrics on the Test Dataset OMOMO~\cite{li2023object}.}
\label{tab:test_metrics}
\resizebox{0.48\textwidth}{!}{
\begin{tabular}{ccccc}
\toprule
Method & Success (\%) $\uparrow$ & $E_{\text{mpkpe}}$ (mm) $\downarrow$ & $E_{\text{mpjae}}$ (rad) $\downarrow$ & $E_{\text{vel}}$ (m/s) $\downarrow$ \\
\midrule
HOVER   & 87.77$_{\pm 0.27}$ & 128.07$_{\pm 3.48}$ & 0.631$_{\pm 0.023}$ & 0.397$_{\pm 0.034}$  \\
ExBody2 & 94.68$_{\pm 0.31}$ & 82.45$_{\pm 3.56}$  & 0.145$_{\pm 0.013}$ & 0.345$_{\pm 0.029}$ \\
OmniH2O & 92.84$_{\pm 0.28}$ & 89.71$_{\pm 4.14}$  & 0.148$_{\pm 0.015}$ & 0.367$_{\pm 0.038}$ \\
TWIST2  & 94.59$_{\pm 0.34}$ & 78.39$_{\pm 2.56}$  & 0.143$_{\pm 0.009}$ & \textbf{0.329$_{\pm 0.015}$} \\
Ours ($\pi^r$) & 95.98$_{\pm 0.33}$ & \textbf{71.52$_{\pm 3.79}$} & \textbf{0.139$_{\pm 0.018}$} & 0.341$_{\pm 0.028}$ \\
Ours ($\pi^h$) & 95.23$_{\pm 0.46}$ & 72.21$_{\pm 4.08}$ & 0.140$_{\pm 0.016}$ & 0.339$_{\pm 0.034}$ \\
Ours ($\pi^e$) & \textbf{98.42$_{\pm 0.29}$} & 75.52$_{\pm 4.29}$ & 0.142$_{\pm 0.019}$ & 0.337$_{\pm 0.032}$ \\
\bottomrule
\end{tabular}
}
\end{table}

%% file: table/table_ablation.tex
\begin{table}[htbp]
\centering
\caption{Ablation study on the curriculum learning (CL) strategy and network size. $*^L$ Denotes the use of a large size network.}
\label{tab:ablation}
\resizebox{0.48\textwidth}{!}{
\begin{tabular}{cccccc}
\toprule
\multirow{2}{*}{Method} & \multicolumn{3}{c}{Tracking Error} & Completion & Perf \\
                        & $E_{\text{mpkpe}}$ (mm) $\downarrow$ & $E_{\text{mpjae}}$ (rad) $\downarrow$ & $E_{\text{vel}}$ (m/s) $\downarrow$ & Success (\%) $\uparrow$ & time$\uparrow$ \\
\midrule
w/o CL                  & 59.01$_{\pm 1.52}$ & 0.126$_{\pm 0.013}$ & 0.348$_{\pm 0.019}$ & 98.17$_{\pm 0.08}$ & \textbf{1.01} \\
w/o CL$^L$              & 54.28$_{\pm 1.65}$ & \textbf{0.121$_{\pm 0.011}$} & \textbf{0.305$_{\pm 0.023}$} & 98.51$_{\pm 0.11}$ & 0.73 \\
w CL                    & 55.27$_{\pm 2.12}$ & 0.125$_{\pm 0.012}$ & 0.327$_{\pm 0.017}$ & 98.85$_{\pm 0.13}$ & 1.00 \\
w CL$^L$                & \textbf{54.22$_{\pm 1.87}$} & 0.123$_{\pm 0.009}$ & 0.320$_{\pm 0.014}$ & \textbf{99.01$_{\pm 0.09}$} & 0.72 \\
\bottomrule
\end{tabular}
}
\end{table}

%% file: table/table_ablation_network.tex
\begin{table}[htbp]
\centering
\caption{Ablation study on the effects of dataset size and network size on the test dataset OMOMO~\cite{li2023object}.}
\label{tab:ablation_dataset}
\resizebox{0.48\textwidth}{!}{
\begin{tabular}{ccccc}
\toprule
\multirow{2}{*}{Method} & \multicolumn{3}{c}{Tracking Error} & Completion \\
                        & $E_{\text{mpkpe}}$ (mm) $\downarrow$ & $E_{\text{mpjae}}$ (rad) $\downarrow$ & $E_{\text{vel}}$ (m/s) $\downarrow$ & Success (\%) $\uparrow$ \\
\midrule
mini                  & 101.36$_{\pm 4.42}$ & 0.203$_{\pm 0.018}$ & 0.590$_{\pm 0.043}$ & 85.13$_{\pm 0.38}$ \\
mini$^L$              & 89.72$_{\pm 3.35}$  & 0.173$_{\pm 0.016}$ & 0.419$_{\pm 0.035}$ & 91.12$_{\pm 0.31}$ \\
full                  & 84.70$_{\pm 2.46}$  & 0.151$_{\pm 0.014}$ & 0.335$_{\pm 0.018}$ & 94.05$_{\pm 0.25}$ \\
full$^L$              & \textbf{80.90$_{\pm 2.87}$} & \textbf{0.145$_{\pm 0.012}$} & \textbf{0.320$_{\pm 0.013}$} & \textbf{94.28$_{\pm 0.25}$} \\
\bottomrule
\end{tabular}
}
\end{table}

%% file: section/S5_conclusion.tex
\section{Conclusion and Future Work}

We presented \oursmethod, a multi-modal and scalable whole-body control framework for humanoid robots. 
By employing modality-specific encoders with a shared latent space, the framework enables unified representation learning for heterogeneous motion reference. 
We analyzed the impact of different modalities on whole-body control and demonstrated strong out-of-distribution generalization. 
Future work will investigate the integration of high-level perception and planning with the learned versatile controller to form a complete hierarchical control framework.